\documentclass{article} 
\usepackage{iclr2022_conference,times}

\usepackage{amsmath,amsfonts,bm}

\def\eqref#1{equation~\ref{#1}}

\def\1{\bm{1}}

\DeclareMathAlphabet{\mathsfit}{\encodingdefault}{\sfdefault}{m}{sl}
\SetMathAlphabet{\mathsfit}{bold}{\encodingdefault}{\sfdefault}{bx}{n}

\usepackage{hyperref}
\usepackage{url}
\usepackage{xcolor}
\usepackage{graphicx}

\usepackage{booktabs}       
\usepackage{amsfonts}       
\usepackage{nicefrac}       
\usepackage{microtype}      
\usepackage{hyperref}
\usepackage{multirow, makecell}
\usepackage[font=small,labelfont=bf]{caption} 

\usepackage{todonotes}
\usepackage{enumitem}

\title{\LARGE SLAM: A Unified Encoder for Speech and \\
Language Modeling via Speech-Text Joint \\
Pre-Training}

\author{Ankur Bapna$^{*1}$, Yu-An Chung$^{*1,2}$, Nan Wu$^{1,3}$, Anmol Gulati$^{1}$, Ye Jia$^{1}$, \\\bf
Jonathan H. Clark$^{1}$, Melvin Johnson$^{1}$, Jason Riesa$^{1}$, Alexis Conneau$^{*1}$, Yu Zhang$^{*1}$ \\
$^{1}$Google Research\\
$^{2}$MIT Computer Science and Artificial Intelligence Laboratory\\
$^{3}$Center for Data Science, New York University\\
}

\makeatletter
\def\nomarkfootnote{\xdef\@thefnmark{}\@footnotetext}
\makeatother

\iclrfinalcopy 
\begin{document}

\maketitle

\begin{abstract}
Unsupervised pre-training is now the predominant approach for both text and speech understanding. Self-attention models pre-trained on large amounts of unannotated data have been hugely successful when fine-tuned on downstream tasks from a variety of domains and languages. This paper takes the universality of unsupervised language pre-training one step further, by unifying speech and text pre-training within a single model. We build a single encoder with the BERT objective on unlabeled text together with the w2v-BERT objective on unlabeled speech.
To further align our model representations across modalities, we leverage alignment losses, specifically Translation Language Modeling (TLM) and Speech Text Matching (STM) that make use of supervised speech-text recognition data.
We demonstrate that incorporating both speech and text data during pre-training can significantly improve downstream quality on CoVoST~2 speech translation, by around 1 BLEU compared to single-modality pre-trained models, while retaining close to SotA performance on LibriSpeech and SpeechStew ASR tasks.
On four GLUE tasks and text-normalization, we observe evidence of capacity limitations and interference between the two modalities, leading to degraded performance compared to an equivalent text-only model, while still being competitive with BERT.
Through extensive empirical analysis we also demonstrate the importance of the choice of objective function for speech pre-training, and the beneficial effect of adding additional supervised signals on the quality of the learned representations.
\end{abstract}

\nomarkfootnote{$^*$Equal contribution. Correspondence to \texttt{\{ankurbpn, aconneau, ngyuzh\}@google.com}.}

\section{Introduction}
Self-supervised learning of text and speech representations has been particularly impactful in natural language processing and speech processing. Since GPT~\citep{radford2018improving},  BERT~\citep{devlin2019bert} and their variations \citep{yang2019xlnet,conneau2019cross,lewis2020bart,raffel2020exploring,Joshi2020SpanBERTIP}, performance on natural language understanding downstream tasks~\citep{socher2013recursive,rajpurkar2016squad,agirre2007semantic,williams2018broad} and monolingual (e.g., GLUE~\citep{wang2019glue}, SuperGLUE~\citep{wang2019superglue}) and multilingual (e.g., XTREME~\citep{hu2020xtreme}, XTREME-R~\citep{ruder2021xtreme}) benchmarks has largely improved thanks to evolving pre-trained models, which leverage increasing amounts of unannotated data~\citep{radford2019language,liu2019roberta,conneau2019unsupervised,wenzek2020ccnet,xue-etal-2021-mt5} and increased model capacity ~\citep{brown2020language,xue-etal-2021-mt5,lepikhin2020gshard,fedus2021switch}. Similarly for speech, unsupervised pre-training has emerged as a predominant approach. Wav2vec 2.0
~\citep{baevski2020wav2vec} and newer variants~\citep{zhang2020pushing} initially showed the strength of pre-training on speech recognition~\citep{panayotov2015librispeech,librilight,zhang2021bigssl} on multiple domains~\citep{hsu2021robust} and languages~\citep{conneau2020unsupervised}.

Self-supervised learning methods in language understanding are designed to be used universally, i.e. a single large pre-trained model for all domains and languages. One big advantage of these universal models is the ability to leverage data skew across domains, tasks and languages; the availability of task or domain-specific data in one language can boost model performance for several languages that the model was pre-trained on. Extending this generalization capability across modalities by having neural networks understand both text and speech at the same time is a natural next step.

Jointly pre-training models on speech and text is a natural choice for multimodal self-supervised learning, given the similarities between the two modalities and the abundance of unannotated text data compared to speech. Recent work has also shown that self-supervised speech representations can be aligned to text with little to no supervision \citep{baevski2021unsupervised}, suggesting the possibility of learning both modalities within a single neural network. However, past work in multilingual modeling in particular has demonstrated the difficulty of learning representations of different data structures, however similar, within a shared network, exposing the so-called transfer interference problem \citep{arivazhagan2019massively}. We show in this work that this trade-off also applies to joint speech-text self-supervised learning.




We study a new multimodal speech-text pre-training approach that leverages data from one modality to improve representations of the other, but also suffers from transfer interference and capacity dilution. Our \emph{Speech and LAnguage Model (SLAM)} consists of a single Conformer~\citep{gulati2020conformer} trained with the SpanBERT objective for text~\citep{Joshi2020SpanBERTIP} and the w2v-BERT~\citep{chung2021w2v} objective for speech. We show that a model using only self-supervised objectives leads to good performance on both modalities, but is outperformed by mono-modal pre-trained models, suffering from significant transfer interference. To reduce the gap, we leverage supervised alignment losses, specifically a translation language model ~\citep{conneau2019cross,zheng2021fused} and speech-text matching~\citep{li2021align} loss. We train our model in a multi-task fashion with the self-supervised and alignment losses. This leads to performance competitive with the state-of-the-art on SpeechStew and LibriSpeech ASR and on CoVoST~2 speech translation tasks. On speech translation, we demonstrate further quality improvements by continuing pre-training on speech-only, outperforming previous approaches by 1 BLEU on average. On text tasks, our joint model loses quality compared to equivalent mono-modal pre-trained models, but remains competitive with initial BERT results~\citep{devlin2019bert}, demonstrating the capacity limitations with modeling two high-resource modalities simultaneously. To the best of our knowledge, our work is the first to study and underline the benefits and limitations of speech-text unsupervised pre-training over mono-modal models, on various speech and text downstream tasks. Our initial results set a new challenge in multimodal self-supervised language understanding.




\section{Related Work}
Self-supervised learning of language representations using neural networks has a long history. In the deep learning era, word2vec~\citep{mikolov2013distributed} initially trained word representations from unannotated data using noise contrastive estimation~\citep{gutmann2012noise,mnih2012fast}. Word2vec was followed by a series of papers that expanded the approach to contextual representations of sentences, including ELMo ~\citep{peters2018deep}, GPT~\citep{radford2018improving}, BERT~\citep{devlin2019bert} and T5~\citep{raffel2019exploring}. They rely on either generative language modelling~\citep{bengio2003neural} or masked language modeling~\citep{taylor1953cloze} (MLM) and these self-supervised pre-training approaches have led to significant improvements on a wide variety of downstream tasks~\citep{wang2019glue, hu2020xtreme}.

In parallel, similar approaches were explored in speech understanding. \cite{chung2016audio} follows the word2vec approach to learn vector representations of variable-length audio segments. \cite{oord2018representation} introduces contrastive predictive coding (CPC) which leverages language modeling and negative sampling to learn speech representations. The first wav2vec model~\citep{schneider2019wav2vec} closely follows this architecture using a noise contrastive binary classification task for unsupervised pre-training. vq-wav2vec~\citep{baevski2020vqwav2vec} proposes to add a vector quantizer similar to VQ-VAE~\citep{oord2018neural}, using Gumbel softmax~\citep{jang2016categorical} or online k-means clustering to quantize the dense speech representations~\citep{eloff2019unsupervised}. When quantized, speech utterances become sequences of discrete tokens belonging to a fixed vocabulary, similar to text, on which BERT is applied. wav2vec 2.0 merges those two separate steps (quantization and contrastive learning) into a unified end-to-end learning procedure that pre-trains a Transformer model. They show significant gains on LibriSpeech \citep{panayotov2015librispeech} as well as on few-shot learning for low-resource languages \citep{conneau2020unsupervised}. w2v-BERT \citep{chung2021w2v} expands wav2vec 2.0 by combining contrastive learning and MLM. \cite{zhang2020pushing} and BigSSL \citep{zhang2021bigssl} explore the limits of large-scale semi-supervised learning with Conformers \citep{gulati2020conformer}.

One approach to utilize data across modalities could involve synthetically transforming the modality of the data; one example being \citet{chen2021injecting} where the authors utilize text-to-speech (TTS) to transform text data into speech, and utilize it for monomodal speech pre-training. Recent advances in self-supervised learning for text, speech and images have led to a new frontier: multimodal self-supervised learning, where a single model learns representations of all modalities using both unannotated and aligned data. VATT Transformer \citep{akbari2021vatt} leverages datasets of more than 100M video-audio-text triplets to learn representations on all modalities at once with noise contrastive estimation. \cite{li2021align} jointly learns to do masked language modeling on text as well as matching image representations to text with parallel data through alignment losses. \citet{jia2021png} learns language representation for text-to-speech synthesis by jointly training on phoneme and grapheme representations with MLM. Perhaps most similar to our work, \citet{zheng2021fused} learn joint speech-text representations by adapting a translation language modeling (TLM) loss \citep{conneau2019cross} to the speech-text setting and studies downstream effect on speech translation.

This work investigates the possibility of developing truly multimodal pre-trained models building on state-of-the-art speech and text pre-training approaches, and highlights the advantages and challenges associated with multimodal pre-trained models by evaluating on a variety of speech and text downstream tasks.

\section{Method}
In this section, we describe each component of our speech-text pre-training framework, SLAM, starting with the model architecture in Section~\ref{sec:model}.
We then present the pre-training objectives and our multi-stage pre-training strategy in Sections~\ref{sec:pretraining} and~\ref{sec:multistage}, followed by introducing the pre-training data in Section~\ref{sec:data}.
Figure~\ref{fig:model} illustrates the overall pre-training framework.

\subsection{Model Architecture}
\label{sec:model}
Our model contains a speech encoder, a text encoder, and a multimodal encoder.
At a high level, the speech and text encoders take speech and text signals as input respectively and extract latent features from them. The latent features from the two modalities are then 
fed to the same multimodal encoder for learning speech-text joint representations. Next, we describe each of these components.

\begin{figure}
\centering
\includegraphics[width=1.0\linewidth]{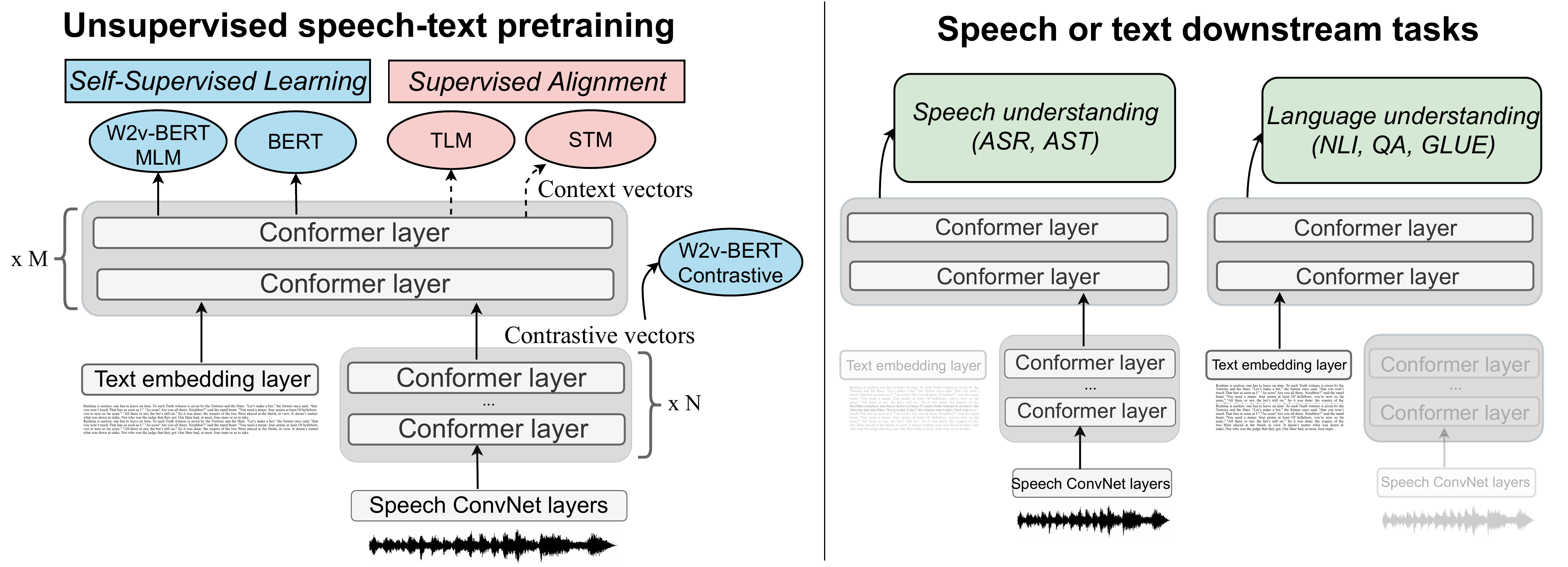}
\captionof{figure}{\textbf{(Left)} Our model consists of a text embedding layer and a speech-specific stack similar to w2v-BERT, the latter consisting of a ConvNet and a series of $N$ Conformer layers. Both the text and speech output embeddings are fed to a series of $N$ shared Conformer layers. Our unsupervised speech-text pre-training approach consists of self-supervised learning objectives (in blue), including w2v-BERT masked language modeling and contrastive losses, as well as the text BERT objective. This can be combined with supervised alignment losses (in red) which leverage speech-text annotated pairs. We leverage in particular the MLM variant of translation language modeling (TLM) and the ranking loss of speech-text matching (STM). \textbf{(Right)} Once pre-trained, the speech part of the shared architecture can be fine-tuned on speech understanding datasets like recognition or translation. The text part of the architecture can be fine-tuned on language understanding tasks.}
\label{fig:model}
\end{figure}

\paragraph{Speech Encoder}
The speech encoder is composed of a convolutional feature encoder followed by a stack of Conformer layers~\citep{gulati2020conformer}, each of which is a series of multi-headed self attention~\citep{vaswani2017attention}, depth-wise convolutions, and feed-forward layers. Given an acoustic feature sequence~$A = (a_1, a_2, ..., a_N)$~(we use~80-dimensional log Mel spectrograms in this work, i.e.,~$a_{i}\in \mathbb{R}^{80}$), the feature encoder---which consists of two 2D-convolutional layers\footnote{We combine information along the time and the Mel spectrogram frequency dimensions.} both with strides~$(2, 2)$---acts as a sub-sampling block that extracts latent speech representations~$X = (x_1, x_2, ..., x_T)$ with a ~4x reduction in the sequence length from the input~$A$.
A linear projection layer is then applied to map the latent representations' dimensionality to that of the subsequent Conformer stack, which further extracts higher-level contextualized speech representations~$C = (c_1, c_2, ..., c_T)$ from the feature encoder output~$X$. For the Conformer stack, we follow the layout described in \citet{gulati2020conformer}, with a model dimension of $1024$, feedforward hidden dimension of $4096$, convolution kernel size $5$ and $8$ attention heads. Different from \citet{gulati2020conformer}, we use group normalization instead of batch norm in our convolution layers which was empirically found to perform better on multimodal mixed batch training. For all our w2v-BERT experiments we use 8 layers in the speech-specific conformer layer stack ($N=8$ in Figure~\ref{fig:model}). Outputs of the speech encoder are layer normalized before being fed to the multimodal encoder. When training with wav2vec 2.0, we skip the additional Conformer block in the speech encoder.

\paragraph{Text Encoder}
The text encoder is a simple token embedding layer that transforms input text into a sequence of token vector embeddings~$W = (w_1, w_2, ..., w_{T'})$.
\footnote{We evaluated using a deep Transformer or Conformer stack for the text encoder but did not find it empirically useful for speech translation or ASR.}
The textual tokens are combined with sinusoidal positional encodings and layer normalized before being fed to the multimodal encoder. We utilize a SentencePiece model~\citep{kudo-richardson-2018-sentencepiece} with a $32k$ token vocabulary.

\paragraph{Multimodal Encoder}
The multimodal encoder is a deep stack of Conformer layers that can take either just speech, or just text, or concatenated speech-text pairs as input. The Conformer layers used in the multimodal encoder are identical to the ones used in the speech encoder. When training with w2v-BERT we use $M=16$ Conformer layers in the multimodal stack, while we use $M=24$ layers when training with wav2vec 2.0.
Depending on the type of input - i.e. just speech, text or a speech-text pair - the model is tasked to solve different self-supervised pre-training objectives.

\subsection{Pre-Training Objectives}
\label{sec:pretraining}
We pre-train the model with four objectives: SpanBERT~\citep{Joshi2020SpanBERTIP} on unlabeled text, w2v-BERT~\citep{chung2021w2v} on unlabeled speech, Translation Language Modeling~\citep{conneau2019cross,zheng2021fused} on paired speech and text data, 
and Speech-Text Matching~\citep{li2021align} on paired and non-paired speech and text data.

\subsubsection{Self-supervised learning objectives}
We use two self-supervised learning objectives that are trained on unannotated text or speech data.

\textbf{BERT} is the self-supervised learning objective applied to unannotated text input~\citep{devlin2019bert}. It aims to learn contextualized textual representations via solving a masked language modeling~(MLM) task. We mask spans of text as in SpanBERT~\citep{Joshi2020SpanBERTIP}.

\paragraph{w2v-BERT} is the self-supervised learning objective used for pre-training on unannotated speech data~\citep{chung2021w2v}. It combines contrastive learning and MLM, where the former trains the model to discretize continuous speech signals into a finite set of discriminative speech tokens, and the latter trains the model to learn contextualized speech representations via solving a masked prediction task consuming the discretized tokens.

\subsubsection{Alignment losses}
Without the presence of paired data, the only incentive for the model to learn joint representation is the inductive bias of having a shared set of Conformer layers. 
Because this is such a strong assumption, we also leverage alignment losses, which use speech-text paired ASR data to explicitly incentivize the model to share representations within the model. We will see below that this leads to better alignment between the speech and text representations, as indicated by better performance on downstream tasks.

\paragraph{Translation Language Modeling (TLM)} was first introduced to align representations between two languages within a shared Transformer. With TLM, parallel sentences are concatenated and sent to a Transformer MLM which predicts missing words, encouraging the model to leverage context from both input languages. In this work, we concatenate speech utterances with their transcriptions using ASR supervised data, similar to \cite{zheng2021fused}. We then train the model to predict masked text or speech spans with BERT or w2v-BERT, encouraging the use of cross-modal context.

\paragraph{Speech-Text Matching (STM)}
predicts whether a pair of speech and text is positive (matched) or negative
(not matched). We use the multimodal encoder’s output embedding of the [CLS] token as the joint representation of the speech-text pair, and append a fully-connected (FC) layer followed by softmax to predict a two-class probability $p^\text{STM}$. The STM loss is:
$$\mathcal{L}_{\text{STM}} = \mathbb{E}_{(I,T)\sim\mathcal{D}} \left[ \text{H} \left(y^{\text{STM}}, \hspace{4pt} p^{\text{STM}}(I,T) \right) \right]$$
where $y^\text{STM}$ is a 2-dimensional one-hot vector representing the ground-truth label, and H is cross-entropy. The STM objective explicitly trains the model to align speech-text pairs, a signal which self-supervised learning cannot explicitly provide.

\subsection{Implementation Details}
When the input only contains speech, the latent speech features~$X = (x_1, x_2, ..., x_T)$ extracted by the speech encoder are directly fed to the multimodal encoder as input.
The speech branch of the model~(i.e., the speech encoder along with the multimodal encoder) is trained to optimize the w2v-BERT objective. Following \citet{chung2021w2v}, we mask approximately $50\%$ of the speech frames with spans of length $10$.
Analogously, when the input only contains text, the latent textual features~$W = (w_1, w_2, ..., w_{T'})$ extracted by the text encoder are fed to the multimodal encoder as input, and the text branch~(i.e., the text encoder along with the multimodal encoder) is trained to optimize the SpanBERT objective. We mask $15\%$ of text tokens with spans of length $5$.

When the input is a speech-text pair, the latent speech and text representations~$C$ and~$W$ extracted respectively by the speech and text encoders are concatenated, forming a sequence with a total length of~$T + T'$ that is fed to the multimodal encoder as input. The multimodal encoder is then trained to simultaneously predict the masked speech features (as in the w2v-BERT objective) and masked text features (SpanBERT). We use more aggressive masking when using paired data to increase the difficulty-level of the task, and to encourage the multimodal encoder to learn to extract useful features across modalities. We mask a single span consisting of $50\%$ of text tokens, and multiple spans masking out $75\%$ of the speech features when training with paired data.

We train the model simultaneously on all these objectives; at every training step the model is trained on a batch of (i) unlabeled speech, (ii) unlabeled text, and (iii) paired speech and text. The gradients of all objectives are aggregated and used to update the model parameters.

\subsection{Multi-Stage Pre-Training}
\label{sec:multistage}
In practice, we find that pre-training the model simultaneously with unpaired and paired data results in the model overfitting to the relatively small paired dataset. To avoid this we pre-train the model in a multi-stage fashion, where we first pre-train the model just on unpaired text and speech, and then optimize it with unpaired and paired data simultaneously. This multi-stage pre-training approach achieves better downstream performance than optimizing all four losses from scratch.
Concretely, we train on 500k updates with the self-supervised losses, and between 250k and 500k additional steps with the alignment losses. We observed improvements of 0.1 to 0.2 WER on LibriSpeech dev-other and 0.3 average BLEU on CoVoST 2 when using the multi-stage strategy as against training with all losses from scratch. In all models that use TLM and/or STM, we utilize multi-stage pre-training.




\section{Experiments}
We first describe our pre-training and fine-tuning setup, including the speech and text datasets used for pre-training as well as all our downstream tasks. We then present our results, including ablations of our approach and comparisons between multimodal and mono-modal models.

\subsection{Pre-training Data}
\label{sec:data}

\textbf{Libri-light \textit{(speech only)}:} The Libri-light (LL-60k) dataset contains 60k hours of unlabeled speech and is used to pre-train all our Masked Speech Models (MSM). LL-60k is the most widely used large unsupervised speech corpus for various pre-training techniques. Each input speech sequence is constructed by first randomly selecting 32-64 seconds segments from the original utterance. From these segments, a contiguous 32 second region is extracted from a random starting point on-the-fly during MSM pre-training as described in~\citep{zhang2020pushing}

\textbf{LibriLM \textit{(text only)}:} The Librispeech text corpus comprises of nearly 803 million tokens from 40M utterances of filtered text derived from 14.5K Project Gutenberg books~\citep{panayotov2015librispeech}.

\textbf{mC4-En  \textit{(text only)}:} The mC4-En  dataset~\citep{Xue2021mT5AM} consists of multiple terabytes of English text data, mined from CommonCrawl. The dataset is publicly available.\footnote{\scriptsize\url{https://huggingface.co/datasets/mc4}}

\textbf{LibriSpeech \textit{(paired data)}:} We use LibriSpeech~\citep{panayotov2015librispeech} fullset (960h) as paired data for Translation Language Modeling (TLM) and Speech-Text Matching (STM). 

\subsection{Downstream Tasks}
We present results on publicly available, well-benchmarked downstream tasks including speech recogntion, speech translation, text normalization and language understanding. 

\paragraph{Speech translation:} CoVoST 2
\citep{wang2020covost}
is a large-scale multilingual speech translation corpus covering translations from 21 languages into English and from English into 15 languages. This represents the largest open dataset available to date from total volume and language coverage perspective.
Following \cite{wang2021large}, we use four English to X directions, specifically German, Catalan, Arabic and Turkish. To evaluate our pre-trained encoders on speech-translation, we fine-tune it as part of a sequence-to-sequence model with a $4$-layer Transformer decoder. The decoder uses a $384$ embedding dimension, $1536$ feed-forward hidden dimension, $4$ attention heads and a $8192$ token multilingual sub-word vocabulary.

\paragraph{ASR:} SpeechStew has 6 public benchmarks, including LibriSpeech~\citep{panayotov2015librispeech}, AMI~\citep{carletta2005ami}, TEDLIUM~\citep{rousseau2012ted}, Common Voice~\citep{ardila2019common}, Switchboard/Fisher~\citep{cieri2003switchboard} and Wall Street Journal (LDC93S6B, LDC94S13B).  In our experiments, LibriSpeech is the same domain as our pre-training speech data (Libri-light) and others are evaluated as out-of-domain corpora. Following~\citet{chan2021speechstew}, the whole dataset (approx. 5k hours) is used to perform our finetuning experiments. Each dataset used is specific to a certain target data condition, for instance LS-960 closely matches LL-60k, AMI dataset is distinct from the LL-60k condition and it contains speech from two kinds of microphones: ($\mathrm{i}$) independent head microphone (IHM); and ($\mathrm{ii}$) single distant microphone (SDM). SpeechStew is composed of datasets chosen from multiple conditions to create a mixed domain aggregate corpus. Details of its processing are described in~\cite{chan2021speechstew}. To evaluate on ASR, we fine-tune our encoder as a conformer-transducer model, following~\citet{chung2021w2v}.

\paragraph{Language understanding:} We consider four main tasks from the GLUE natural language understanding benchmark: the MNLI natural language inference benchmark~\citep{williams2018broad}, the Quora Question Pair (QQP) classification dataset,\footnote{\scriptsize\url{https://www.kaggle.com/c/quora-question-pairs}} the QNLI question answering task~\citep{wang2019glue} and the SST-2 sentiment analysis dataset~\citep{socher2013recursive}. We report accuracy on the dev sets of each dataset (except SST-2 where we report test accuracy) and compare our results to BERT, SpanBERT and RoBERTa.

\paragraph{Text normalization:} Text normalization --- also referred to as text verbalization --- is a core task in the text-to-speech (TTS) community. Text normalization takes as input raw unverbalized text---as typically found in written form---and produces a verbalized form of that text, expanding numbers, dates, abbreviations, etc. The output of this task is a word-for-word spoken transcript---the input format expected by TTS systems. For example, \textit{``A 1951 Lancia V6''} would become \textit{``a nineteen fifty one lancia v six''} while \textit{``Dial 1951 for 6V batteries''} might become \textit{``dial one nine five one for six volt batteries.''}. We consider the English task data from \cite{sproat-textnorm} and compare our results to those of \cite{stahlberg-kumar-2020-seq2edits}, known to be previously state-of-the-art. We report sentence error-rate on the test set for all our experiments. When evaluating on text normalization, we fine-tune our encoder with a $3$-layer transformer decoder with a model dimension of $512$, hidden dimension of $1024$ and $4$ attention heads. 

\subsection{Main Results}
In this section, we analyze the results of fine-tuning our models on speech and text downstream tasks.

\subsubsection{Speech Translation}
We present our results on CoVoST 2 En-X translation in Table~\ref{t:covost}. We compare our models against results from~\citet{wang2021large}, specifically against fine-tuning wav2vec 2.0 on speech-translation, with and without LM fusion. Our speech-only baselines trained using wav2vec 2.0 improve over ~\citet{wang2021large} by over 1 BLEU, possible due to increased encoder capacity. Our w2v-BERT speech-only encoder further improves performance by around 0.4 BLEU. 

The addition of mC4-En  data to pre-training results in a drop of around 1.3 BLEU for the w2v-conformer, a concrete example of the interference issue. In comparison, the w2v-BERT speech-text model is only worse than its speech-only counterpart by approximately 0.6 BLEU. The addition of alignment losses results in the joint pre-trained model matching the speech-only baseline and alleviates interference.

We continue training our TLM + STM joint model on unlabeled speech-only data to alleviate the capacity limitation in the multimodal pre-trained model. Fine-tuning this speech-adapted model on CoVoST results in a model that outperforms our best speech-only model by almost 1 BLEU point, illustrating positive cross-modal transfer and the advantages of multimodal pre-training.

\begin{table}[t]
\caption{BLEU on CoVoST 2 speech translation comparing our speech-text pre-trained models against speech-only pre-training and pre-existing baselines.}
\centering
\resizebox{0.9\textwidth}{!}{%
\begin{tabular}{llcccccccc}
\toprule
\bfseries \# & Model & \# Params & Text data & En-De & En-Ca & En-Ar & En-Tr & Avg \\
\midrule
& \bfseries Prior Work \\
1 & \quad wav2vec-2.0 \citep{wang2021large} & 300M & - & 23.8 & 32.4 & 17.4 & 15.4 & 22.3 \\
2 & \quad \ wav2vec-2.0 + LM \citep{wang2021large} & - & - & 24.9 & 34.0 & 18.0 & 16.7 & 23.4 \\
\midrule
& \bfseries Our Work Speech-only \\
3 & \quad w2v-conformer & 600M & - & 27.1 & 33.1 & 18.8 & 15.6 & 23.7 \\
4 & \quad w2v-bert & 600M & - & 27.4  & 33.9 & 19.0 & 15.9 & 24.1 \\
\midrule
& \bfseries Our Work Speech-text \\
5 & \quad w2v-conformer + bert  & 600M & mC4-En & 25.4 & 30.5 & 18.5 & 15.2 & 22.4 \\
6 & \quad  w2v-bert + bert (SLAM)  & 600M & mC4-En & 26.9 & 33.1 & 18.1 & 16.1 & 23.5 \\
7 & \quad  SLAM-TLM  & 600M & mC4-En & \bf 27.5 & 33.4 & 18.9 & 16.6 & 24.1 \\
8 & \quad  SLAM-TLM-STM  & 600M & mC4-En & 27.2 & 33.3 & 18.5 & 16.8 & 24.0 \\
\midrule
& \bfseries Our Work Speech-text $\rightarrow$ Speech-only\\
9 & \quad  SLAM-TLM-STM $\rightarrow$ w2v-bert  & 600M & mC4-En & 27.1 & \bf 34.2 & \bf 21.2 & \bf 17.5  &  \bf 25.0 \\
\bottomrule
\end{tabular}}
\label{t:covost}
\end{table}

\subsubsection{Speech Recognition}

\begin{table}[htbp]
  \caption{WERs~(\%) when using the LibriSpeech 960h as supervised data. For all methods, both self-training and LM fusion are not used. References are where the numbers are quoted from.}
  \vskip 0.1in
  \label{t:librispeech}
  \centering
  \resizebox{0.65\width}{!}{%
  \begin{tabular}{llcccccc}
    \toprule
    \# & \bfseries Model & \#Params & Text data& \bfseries dev & \bfseries dev-other & \bfseries test & \bfseries test-other \\
    \midrule
    & \bfseries Prior Work (no LM) \\
    1 & \quad  wav2vec~2.0-CTC~\citep{baevski2020wav2vec} & 300M & -
    & 2.1 & 4.5 & 2.2 & 4.5\\
    2 & \quad  w2v-Conformer~XL~\citep{zhang2020pushing} & 600M & -
    & 1.7 & 3.5 & 1.7 & 3.5\\
    3 & \quad  w2v-bert~XL~\citep{chung2021w2v} & 600M & -
    & 1.5 & 2.9 & 1.5 & 2.9 \\
    4 & \quad  w2v-bert~XXL~\citep{chung2021w2v}  & 1B & -
    & 1.5 & 2.7 & 1.5 & 2.8 \\    
    \midrule
    & \bfseries Our Work Speech-only \\
    5 & \quad  w2v-bert~XL (Group Norm) & 600M & -
    & 1.5 & 2.9 & 1.6 & 2.9 \\
    & \bfseries Our Work Speech-text \\
    6 & \quad w2v-conformer + bert & 600M & LibriLM
    & 1.7 & 4.0 & 1.9 & 4.0 \\
    7 & \quad w2v-bert + bert (SLAM) & 600M & LibriLM 
    & 1.6 & 3.2 & 1.7 & 3.3 \\
    9 & \quad SLAM-TLM (one stage) & 600M & LibriLM 
    & 1.6 & 3.1 & 1.7 & 3.2 \\    
    8 & \quad SLAM-TLM & 600M & LibriLM
    & 1.5 & 2.9 & 1.6 & 3.1 \\    
    9 & \quad SLAM-TLM-STM & 600M & LibriLM  
    & 1.7 & 3.0 & 1.6 & 3.2 \\    
    10 & \quad SLAM-TLM-STM & 600M & C4
    & 1.7 & 3.2 & 1.7 & 3.2 \\
    \midrule
    & \bfseries Our Work Speech-text $\rightarrow$ Speech-only\\
    11 & \quad SLAM-TLM-STM $\rightarrow$ w2v-bert & 600M & mC4-En  	
    & 1.6 & 3.0 & 1.6 & 3.1 \\
    \bottomrule
  \end{tabular}
  }
\end{table}

\begin{table*}[t]
\caption{WERs (\%) across multiple tasks and settings from the SpeechStew benchmark compared against pre-existing baselines. $^\dagger$We follow \cite{likhomanenko-arxiv-2020} and remove punctuations during evaluation.}
\centering
\resizebox{1.0\textwidth}{!}{%
\begin{tabular}{llcccccccccc}
\toprule
\# & \bfseries Model&\#Params&Text data&\multicolumn{2}{c}{AMI} & Common Voice$^\dagger$ & \multicolumn{2}{c}{Switchboard/Fisher} & TED-LIUM & WSJ & Avg\\
\midrule
&  & & &  IHM & SDM1 & & SWBD & CH & & eval92 & \\
\midrule
& \bfseries Prior Work \\
1 & \quad w2v-conformer \citep{speechstew} & 1B & - & 9.5 & \textbf{22.7} & \textbf{8.4}  & 4.8 & 10.6 & 5.7 & \textbf{1.3} & 9.1 \\
\midrule
& \bfseries Our Work Speech-only \\
2 & \quad w2v-conformer & 600M & - & 9.6 & 23.8 & \textbf{8.4}  & 4.7 & 9.2 & 5.6 & 1.4 & 9.1 \\
3 & \quad w2v-bert & 600M & - & \textbf{9.1} & \textbf{23.1} & 8.6 & \textbf{4.5} & 9.0 & 5.4 & \textbf{1.3} & \textbf{8.7}\\
\midrule
& \bfseries Our Work Speech+text \\
5 & \quad w2v-bert + bert (SLAM) & 600M & LibriLM & 9.4	& 24.3 & 9.7  & 4.8 & 11.0 & 5.9 & 1.5 & 9.5\\
6 & \quad SLAM-TLM & 600M & LibriLM & 9.2 & 23.8 & 8.7  & 4.9 & \textbf{8.9} & 5.8 & 1.3 & 9.3 \\
7 & \quad SLAM-TLM-STM & 600M & LibriLM & 9.3	& 23.5 & 8.6  & 4.6 & 9.1 & 5.6 & \bf 1.3 & 9.0\\
8 & \quad SLAM-TLM & 600M & mC4-En & 9.4 & 24.7 & 8.9	& 4.8 & 9.2	& \bf 5.1 & 1.5 & 9.1 \\
9 & \quad SLAM-TLM-STM & 600M & mC4-En & 9.5 & 25.3 & 9.0  & 4.6 & 9.0 & 5.3 & 1.5 & 9.2 \\
\bottomrule
\end{tabular}}
\label{t:speechstew}
\end{table*}

In Table~\ref{t:librispeech}, we present our results on the Librispeech 960h ASR benchmark.
We compare our unified speech-text encoders to a number of state-of-the-art self-supervised representation learning methods from the literature, including wav2vec~2.0~\citep{baevski2020wav2vec} and w2v-bert~\citep{chung2021w2v}. 

As shown in Table~\ref{t:librispeech}, w2v-BERT is consistently better than w2v-Conformer with the text modality by 17\% relative (line 6 and 7). However, simply adding the text modality with LibriLM data hurts ASR performance by ~14\% relative compared to the speech-only model, from 2.9 to 3.3 average WER (line 5 and 7) on testother.
By adding TLM loss (line 8), we are able to reduce the interference and we bridge most of this gap, matching performance on dev/devother/test, and only 0.2\% worse on testother compared to the mono-modal model. We conclude that the alignment losses help the model align the two modalities, resulting in better use of shared parameters and reduction in the interference between the speech and text modalities. Further introducing STM loss does not improve ASR performance (line 9), but it still performs better than the model without alignment losses.
As we increase the amount of text data from LibriLM to mC4-En (line 10), we observed a regression on devother and testother. We conclude that the model needs more capacity to learn from the out-of-domain and larger text dataset. Similar to speech translation, if we further pre-train the model with speech only data, there is 0.1\% consistent improvement over all the test sets (line 10 and 11). 

In Table~\ref{t:speechstew}, we present our results on 5 ASR benchmarks using SpeechStew supervised data. Note that the unified encoder model has not seen any paired data during pre-training on these out-of-domain benchmarks. We notice that the alignment losses still improve over the baseline multimodal model (line 5 to 7). Interestingly, mC4-En data improves performance on TED-LIUM but is worse on AMI compared to pre-training on LibriLM.\footnote{TED-LIUM is clean speech from the TED-talks domain and thus likely to benefit from more text data, whereas AMI has natural speech from meetings and might benefit from additional capacity devoted to acoustic modeling.}

\subsubsection{Natural Language Understanding}
We report results on four natural language understanding tasks from GLUE in Table~\ref{t:glue}. We compare our methods to the original BERT model of \cite{devlin2019bert} and its extended versions, SpanBERT~\citep{Joshi2020SpanBERTIP} and RoBERTa~\citep{liu2019roberta} which are trained on comparable objectives and comparable text data respectively. We report dev results for MNLI, QNLI and QQP, as test sets are not available for these tasks. We see that our SpanBERT-conformer text-only baseline obtains competitive results with SpanBERT but is outperformed by RoBERTa, possibly because of the Conformer architecture and the optimized pre-training and fine-tuning of the RoBERTa architecture. Doing an apples-to-apples comparison of our text-only model and our speech-text architectures, we observe significant decrease in performance when adding the speech modality. On MNLI, for instance, we go from 87.9\% accuracy (line 5) down to 83.3\% accuracy with our full model (line 8), or from 95.4 on SST-2 to 93.9\%. We observe some gains in performance when using alignment losses over the fully self-supervised learning approach (line 6) which only slightly alleviates the interference problem. Given the large amount of data in both speech and text for English, it is likely that the capacity of the model is a limiting factor for understanding both modalities simultaneously. We believe that alleviating capacity limitations by inducing better cross-modal alignments is an important challenge. We leave the investigation of larger-capacity models and lower-resource languages for future work.

\subsubsection{Text Normalization}
In addition to GLUE, we also evaluate and report sentence error rate for text normalization and compare our approach to \cite{stahlberg-kumar-2020-seq2edits} in Table~\ref{t:glue}. Our baseline text-only model improves over the previous state-of-the-art by $0.25\%$ absolute (lines 5 and 6). Adding speech during pre-training results in worse performance compared to our text-only pre-training, but the addition of TLM and STM alignment losses is able to recover some of the lost quality (lines 6 to 8). Based on this, we suspect that future work in cross-modality alignment may yield improvements on this task.

\begin{table*}[t]
\caption{Performance on four GLUE tasks and text-normalization against text-only baselines. We report dev accuracy scores for MNLI, QNLI and QQP, test accuracy for SST-2 and test sentence-error-rate for Text-Norm.}
\centering
\resizebox{0.95\textwidth}{!}{%
\begin{tabular}{llcccccccc}
\toprule
\# & \bfseries Model & \# Params & Text data & MNLI & QNLI & QQP & SST-2 & Text-Norm \\
\midrule
& \bfseries Prior Work \\
 1 &\quad BERT \citep{devlin2019bert} & 340M & Wiki+Books & 86.6 & 92.3 & 91.3 & 93.2 & - \\
2 & \quad SpanBERT \citep{Joshi2020SpanBERTIP} & 340M & Wiki+Books & 87.0 & 93.3 & - & 94.8 & - \\
3 & \quad RoBERTa \citep{liu2019roberta} & 340M & CC & 90.2 & 94.7 & 92.2 & 96.4 & - \\
4 & \quad Seq2Edits \citep{stahlberg-kumar-2020-seq2edits} & - & - & - & - & - & - & 1.36 \\
\midrule
& \bfseries Our Work Text-only \\
5 & \quad SpanBERT-conformer & 450M & mC4-En & 87.9 & 92.6 & 91.8 & 95.4 & 1.11 \\
\midrule
& \bfseries Our Work Speech-text \\
6 & \quad w2v-bert + bert (SLAM) & 450M & mC4-En & 82.3 & 88.4 & 90.6 & 94.2 & 1.30\\
7 & \quad SLAM-TLM & 450M & mC4-En & 83.6 & 90.1 & 91.0 & 94.3 & 1.28 \\
8 & \quad SLAM-TLM-STM  & 450M & mC4-En & 83.3 & 90.0 & 91.0 & 93.9 & 1.19 \\
\bottomrule
\end{tabular}}
\label{t:glue}
\end{table*}

%



\section{Discussion}
In this work, we demonstrate that a single encoder model can be pre-trained to learn strong contextualized representations of speech and text simultaneously. We combine self-supervised learning objectives for text (BERT) and self-supervised approaches for speech (w2v-BERT) to learn a joint Speech and LAnguage Model (SLAM). Downstream evaluations on speech and language understanding tasks, including LibriSpeech and SpeechStew ASR, CoVoST 2 speech translation, four GLUE tasks, and text-normalization uncover significant interference challenges when pre-training simultaneously on high-resource modalities. Using alignment losses such as translation language modeling and speech-text matching which leverage speech-text supervised aligned data, we show that we can improve the cross-modal representation alignment and improve over mono-modal models on the speech translation tasks, while maintaining state-of-the-art performance on speech recognition.
We hope that this work would motivate further research on extending the universality of self-supervised learning of language representations to the multimodal speech-text setting.


\newpage
\bibliography{iclr2022_conference}
\bibliographystyle{iclr2022_conference}

\appendix

\end{document}